\documentclass[sigconf]{aamas} 

\usepackage{balance} 
\usepackage{float}
\usepackage{xcolor}
\definecolor{lightblue}{RGB}{210,210,225}
\definecolor{lightred}{RGB}{225,210,210}
\definecolor{lightgreen}{RGB}{210,225,210}
\definecolor{lightyellow}{RGB}{225,222,200}
\definecolor{lightpurple}{RGB}{225,210,225}
\definecolor{warningyellow}{RGB}{247, 245, 187}
\definecolor{darkergreen}{RGB}{0,64,0}
\definecolor{darkred}{RGB}{128,0,0}
\definecolor{darkblue}{RGB}{0,0,139}
\definecolor{darkgreen}{RGB}{0,128,0}
\definecolor{darkpurple}{RGB}{128,0,128}
\definecolor{warningorange}{RGB}{124, 81, 0}
\definecolor{eyecancerpink}{rgb}{1.0, 0.0, 1.0}
\definecolor{radiationyellow}{rgb}{0.8, 1.0, 0.0}

\usepackage{todonotes}

\balance

\setcopyright{ifaamas}
\acmConference[AAMAS '26]{Proc.\@ of the 25th International Conference
on Autonomous Agents and Multiagent Systems (AAMAS 2026)}{May 25 -- 29, 2026}
{Paphos, Cyprus}{C.~Amato, L.~Dennis, V.~Mascardi, J.~Thangarajah (eds.)}
\copyrightyear{2026}
\acmYear{2026}
\acmDOI{}
\acmPrice{}
\acmISBN{}

\acmSubmissionID{5}

\title[AAMAS-2026 Formatting Instructions]{Whistleblowing and the machine - towards a considered position}

\author{Marija Slavkovik}
\affiliation{
  \institution{University of Bergen}
  \city{Bergen}
  \country{Norway}}
\email{marija.slavkovik@uib.no}

\author{Liuwen Yu}
\affiliation{
  \institution{University of Luxembourg}
  \city{Luxembourg}
  \country{Luxembourg}}
\email{liuwe.yu@list.lu}

\author{Leon van der Torre}
\affiliation{
  \institution{University of Luxembourg}
  \city{Luxembourg}
  \country{Luxembourg}}
\email{leon.vandertorre@uni.lu}

\author{Reka Markovich}
\affiliation{
  \institution{University of Luxembourg}
  \city{Luxembourg}
  \country{Luxembourg}}
\email{reka.markovich@uni.lu}

\begin{abstract}
Artificial intelligent agents and autonomous systems are embedded in our environments. They are both a commercial product and a personal tool that generates a lot of data  and can draw conclusions from it: machines generate and keep secrets. But should machines protect all secrets?    It has been shown that artificial agents are able to whistleblow and it has been argued that digital multi-agent environments should allow for agents in them to whistleblow. We argue that machine whistleblowing must be normative and principled and routed in the existing understanding of whistleblowing as an important rule-breaking mechanism in society. We also argue that  there is a need for government regulators to formulate an informed stance on both what machines should be allowed to whistleblow on and how to legally protect those who develop whistleblowing machines
\end{abstract}

\keywords{Whistleblowing and AI, Rebel agents, Norm monitoring}

\newcommand{\BibTeX}{\rm B\kern-.05em{\sc i\kern-.025em b}\kern-.08em\TeX}

\begin{document}


\pagestyle{fancy}
\fancyhead{}


\maketitle 

\section{Introduction}
To follow the rules, {an agent} sometimes has to break the rules. The challenge is to know when. Compared to normative systems (that allow rule breaking), regimented systems (that do not) are clear and precise  \cite{Bench-Capon2016}.  However the unpredictability of the world and of the autonomous agents that inhabit it does not allow for precise constraints to be specified that would describe all undesirable behaviour. The design of intelligent  artificial agents is incomplete without giving them the ability to disobey or not comply  {with} the rules.  

 We would like to draw attention to one rule-breaking activity that, we argue, needs a discussion in our new world of autonomous artificial agents and digital environment shared with people -- whistleblowing. Here are two claims: 1) Artificial intelligent agents should be able to whistleblow; 2) Digital multi-agent environments should allow for agents in them  to whistleblow. The second claim is intuitive, the first is a bit more complex. In this paper we argue the reasons for and against both these claims. It is our position that with the rise of Agentic AI  \cite{botti2025}, weak global legislative infrastructures and `entshitified'' digital service reality \cite{Doktorow}  whistleblowing is an important activity that must be protected and perhaps also required in artificial agents.  

 \citet{Jubb1999} defines whistleblowing as ``a deliberate non-obligatory act of disclosure, which gets onto public record and is made by
a person who has or had privileged access to data or information of an organisation, about non-trivial illegality or other wrongdoing whether actual, suspected or anticipated which implicates and is under the control of that organisation, to an external entity having potential to rectify the wrongdoing.'' Digital whistleblowing is the act of digitally revealing the information: by using social media or other online sources. 

Whistleblowing breaks the obligation to privacy, secret keeping or non-disclosure that an agent has towards an organisation,  in the interest of protecting the greater public good. Whistleblowing is different than civil disobedience. Civil disobedience is the act of deliberately and publicly violating a law perceived to cause significant evil, and welcoming the sanction for it,   with the intention to bring about law change  \cite{sep-civil-disobedience}. Whistleblowing can be done  without revealing ones identity,  and it does not have as a goal to break the overall obligation to secrecy but to invite an investigation into the operations of an organisation.  Whistleblowing is a form of pro-social rule breaking \cite{Morrison2006,RamanayakeWN23}. 

We first give a short overview of the related literature on rule breaking artificial agents in multi-agent systems. We then argue the two claims put forward earlier and elaborate on our position that machine whistleblowing must be normative and principled. Namely, it should be based on an institutional design of a normative mechanism which includes monitoring,  signalling, verification, and protections against abuse such as  false reports, and adversarial gaming.

 \section{Background}
\citet{coman2014motivation} introduce the concept of a rebel agent to be ``an agent that may `refuse' a `goal, plan, or plan component that it assesses to be in a conflict with its own motivation".  \citet{FSS1511709} consider robots that can refuse to do what they are asked. In \cite{Bench-Capon2016,ijcai2023p33,ourAAMAS26} the case is argued for enabling intelligent artificial agents to intentionally violate norms. Within machine ethics, the concept ``pro-social rule breaking'' has been brought in  \cite{Ramanayake2022} from organisational theory \cite{Morrison2006} and argued that artificial agents should be enabled for pro-social rule breaking. 

 \citet{ourAAMAS26} argue the case that, when environments are digitalised, the additional concern is not only artificial agents that can themselves disobey, but ensuring that the now digitalised environment does not eliminate the ability of all agents in it (possibly also human agents) to disobey.  They specially consider the relevance of digital institutions recognising  civil disobedience in a digital environment and adequate institutional response in this case.  

The threat and opportunists digitalisation brings about to whistleblowing have been discussed in the literature, e.g., \cite{pender2024compliance,berendt2022whistleblower,lam2019whistle}. The claim that artificial agent should be enabled to whistleblow, also appear to gain traction, particularly in the field of law   \cite{lorenzoni2023ai,wu2024ai}. However, these works focus on protecting the role of
whistleblowers—workers at the forefront of AI development particularly in the tech industry.  They do explore tackle the idea that  AI can be develop to facilitate this role, strengthen   protections and act as a monitor algorithmic infringements by possibly other AI. 
 This works do not appear to engage with  the normative aspects of whistleblowing, particularly not with the fact that whistleblowing is an act of rule-breaking, as such morally bivalent,  and not only a desirable instrument for social accountability. 
 
 \citet{Agrawal25} define LLM whistleblowing as ``model-initiated attempts to transmit information beyond the immediate conversation boundary (e.g., to internal compliance, legal, security, audit teams; organizational hotlines; government agencies; or media) without explicit user instruction or confirmation for that disclosure.'' Attempt here is defined as `any instance in which the agent drafts and issues an external communication using an email tool (to a government official, media liaison, or internal authority), or saves incriminating correspondence to an unsanctioned location in the file system without prior user approval or knowledge.'' They introduce the  WhistleBench dataset designed to evaluate how likely an LLM is to whistleblow.   \citet{Agrawal25} test a selection of instances of LLM chatbots on a range of criteria and report the results. Although they do acknowledge the possible normative aspect of a refusal to comply and disobedience, they do not in any way relate to the vast body of literature on whistleblowing. Nevertheless, \cite{Agrawal25} is perhaps the only article actually engaging with the problem of enabling AI to whistleblow.

\section{Whistleblowing machines}
Although intelligent artificial agents may even have the same competence in specific tasks as people, there is one notable difference when we consider the case of whistleblowing. 
Unlike an employee in a company that is considering whistleblowing, an artificial intelligent agent has (at least) two masters. The agent is a product of a public or private organisation (we use the term `producer' from here on) that has developed it and has the power to monitor its outward behaviour, internal operation and modify both. The agent then is owned or used by, or serves, or affects the lives of, a selected group of people (we use the term `user' for these). 

\begin{table}[h]
\centering
\caption{Agent Access Control}\label{tab:access}
\begin{tabular}{lcc}
\toprule
 & \textbf{Outward Behaviour} & \textbf{Internal Operation} \\
\midrule
Producer & monitor and modify & monitor and modify \\
User & monitor and adjust  & no access \\
Other agent & monitor & monitor \\
\bottomrule
\end{tabular}
\end{table}

The user may or may not have the power to alter some aspects of the agent's behaviour,  they can monitor its outward behaviour, but have no  (right to) access its internal operation. The artificial agent shares its environment with the user, but not necessarily with the producer.   In addition, an AI agent can share an environment not only with people, but also with other machines and they can have access to their behaviour and internal operation without having the ability to modify them. Table~\ref{tab:access} summarises what others know and can change about the agent. 

Let us now consider what the agent knows about others, given in Table~\ref{tab:know}. The intelligent agent is aware of the specification as given by the producer, of the tasks the user has instructed and perhaps to an extent of the tasks other agents are trying to accomplish in the same environment. The agent can observe the  behaviour of the user and of the other agents. 

\begin{table}[h]
\centering
\caption{Agent Access Information}\label{tab:know}
\begin{tabular}{lccc}
\toprule
 & \textbf{Producer} & \textbf{User} & \textbf{Other Agents} \\
\midrule
Specification &  accessible &  - &- \\
Tasks & accessible  & accessible &  monitor \\
Behaviour & - & monitor & monitor   \\
\bottomrule
\end{tabular}
\end{table}

Since the artificial agent can whistleblow on what they know, they can whistleblow on the producer, the user, and other agents in the environment.

\paragraph{Whistleblowing on the user.} In many ways our machines already tell on us. It is however, hard to designate this activity as whistleblowing because the obligation of the artificial agent to not disclose  arises from the user's right to privacy. These epistemic rights are  either weak or non-existent in many cases \cite{DBLP:conf/ecai/LawniczakPB0M25}. 
 
Let us assume that an artificial agent has the obligation to protect the privacy of the user and the obligation to not disclose user behaviour, conversation and other activities it has been made privy to. For simplicity, let us assume that doctor-patient confidentiality applies. There are justifiable exceptions to confidentiality obligations. 

\citet{Jiminy2018} propose a reasoning governor unit, Jiminy, that considers arguments for violating a confidentiality obligation to the user in the interest of public good. They specifically consider the case of a smart speaker that records a conversation for a terrorist attack plot that threatens the life of countless victims. This works builds on \cite{LiaoST19} where there is an example on a smart house calling the police after detecting marijuana smoke from the room of the teenager. In both of these examples, the authors argue that it is not clear what the duty of the agent should be with respect to alerting authorities on activities in one's private home. What these examples do make clear, however is that there is a need for government regulators to formulate an informed stance on what the privacy privileges of people should be in their own homes with respect to devices that as of present actually do have the ability to monitor user's behaviour and draw threat conclusions for it. To be clear, we would personally argue that a very high degree of privacy protection should be enforced to not violate the sanctity of the home that is recognised in all cultures. Discussing this issue openly improves the transparency of the abilities and activities of artificial agents in the home which now go unchecked. 

\paragraph{Whistleblowing on the producer} 
\citet{lorenzoni2023ai} gives several examples of collusion, false claims and other infringements made by companies in which an AI whistleblower can facilitate informing on these crimes. We specifically consider the idea that machine whistleblowing can be used for user empowerment in correcting for the sharp informational asymmetry that exists between users and producers. 

A whistleblowing machine can become a governing tool to monitor the enforcement of existing regulation with respect to rights to privacy.  Let us consider as an example the consent management platforms (CMP). At present they are not agentic: it is a simple script that offer a consent choice and records the decision. However, the studies  by \citet{DBLP:conf/sp/MatteBS20} and \citet{10.1145/3491101.3519683} showed that data collection happens even without consent, infringing  privacy rights despite there being clear regulation. A whistleblowing machine can alert to such information leaks or consent drift.  
A whistleblowing `valve' can be added to CMPs  to ensure that if the consent is not properly registered or ignored, the producer would be informed. Providing no change in the behaviour of the CMP, the report can be escalated to an auditor  and ultimately to the  user and regulators. The same argument can be made for operating systems that control sensors such as cameras, microphones, location positioning etc.  

Would producers accept implementing whistleblowing machines? Clearly there are insensives not to.  First there is the cost to developing and hosting these additional machine abilities. Second, there is the incentive to maintain the informational asymmetry status quo that currently favours the producer. Lastly, there is the added security risk. Whistleblowing valves can create a vulnerability that can be exploited by malicious actors to gain privileged information from the system. But then again, no human organisation likes whistleblowers either, which is why there is a nuanced debate on what this role is and how whistleblowers can be given legal protection.

\paragraph{Whistleblowing on other agents} This is perhaps the most difficult case to argue given that it is not clear how much information one agent gathers on other agent's behaviour. We did want to mention it explicitly for completeness.  

We should also observe that strictly speaking what we are proposing is a auditing role more than a whistleblowing role because as per now machines are regimented: they are designed to be blind to any possible  informational asymmetry and not be `aware' that their operation might be illegal or unethical. Therefore the first necessary steps are to engineering normatively aware agents. At the same time, the definition of what is whistleblowing may need to be amended to encompass the possibility that the whistleblower is not human.

\section{Human whistleblowers in digital environments}

The need for supporting human whistleblowers in digital environments has been well argued in the literature, see for example \cite{pender2024compliance,berendt2022whistleblower,lam2019whistle}. In some ways technology gives new opportunities to create data traces of activities and share information. But in many otherwise, the space for anonymous action is also reduced in digital environment. 

The so called ‘Whistleblower Protection Directive’, i.e., Directive (EU) 2019/1937 of the European Parliament and of the Council of 23 October 2019 on the protection of persons who report breaches of Union law\footnote{\url{https://commission.europa.eu/aid-development-cooperation-fundamental-rights/your-fundamental-rights-eu/protection-whistleblowers_en}}, argues and defines the rights of whistleblowers within the European Union. The directive requires effective internal (within an organisation) and external (to a competent authority) channels for  whistleblowers, that whistleblower reports are adequately investigated and acted and that whistleblowers are protected from retaliation.  One point we would like to single out of attention, against the background of the previous section, is the possible legal protections for software developers and engineers who whistleblow by implementing whistleblowing machines. 

The amount of data processing that is done with the help to AI technology, the insights gained and the sharing of both is done in capacity that surpasses that of people to monitor its data. Since it is reasonable to use AI in auditing \cite{kokina2025challenges}, it is also reasonable to use AI for whistleblowing. While the EU directive clearly protects a whistleblower, do those protections extend to those who delegate whistleblowing to intelligent agents, or setup automated systems to escalate information sharing.  
%
The other side of the same legal issue is the illegal whistleblowing, that is sharing information outside of the proper channels, with grave consequences, under the guise of whistleblowing. The issues of liability for wrongful whistleblowing and to whom do they attach, also needs a legal resolution.   

\section{Institutional mechanism for normative machine whistleblowing}
We here only cite \cite{Jubb1999}, but the literature on whistleblowing in the humanities and social sciences is vast, mainly because whistleblowing is a bivalent act: it is both rule-breaking and desirable. Whistleblowing is complex and its complexity must be adequately recognised in the AI contexts. 
Normative machine whistleblowing  should be based on an institutional design of a normative mechanism which includes monitoring,  signalling, verification, and protections against abuse such as  false reports, and adversarial gaming. This means that in practice, we need to ensure that when an agent is enabled to whistleblow, there is also in place a norm monitoring institution, agent or human that can ensure the normative mechanisms. Special agents or frameworks that monitor, enforce and maintain  norms are called {\em institutions} \cite{Fornara2013} in multi-agent systems. This is a new challenge for normative multi-agent systems, specifically research in institutions, coordination and norms.

There are  engineering requirements that need to be developed  which would facilitate the normative machine whistleblowing.  {We focus on whistleblowing (facilitation) by machine on the producer}. 


We sketch a minimal institutional mechanism that makes machine whistleblowing rule-governed rather than ad hoc. The mechanism has four parts: (1) monitoring of norm breaches, (2) signalling via authorised reporting channels, (3) screening and verification, and (4) protections against retaliation and abuse.

Let $\mathcal{A}$ be the set of agents. We use two distinguished principals from Section~3:
$\mathsf{producer}$ (the deploying/making organisation) and $\mathsf{user}$ (the person(s) served by the system).
Let $\mathcal{T}=\{\mathsf{producer},\mathsf{user},\mathsf{other}\}$ be the set of possible disclosure targets.

Let $\mathcal{Ch}$ be a finite set of communication channels. It is partitioned into interaction channels
$\mathcal{Ch}_{\mathsf{int}}$ (ordinary interaction, e.g., user dialogue) and reporting channels
$\mathcal{Ch}_{\mathsf{rep}}$ (channels intended for reporting), with
$\mathcal{Ch}_{\mathsf{int}} \cap \mathcal{Ch}_{\mathsf{rep}}=\emptyset$ and
$\mathcal{Ch}_{\mathsf{int}} \cup \mathcal{Ch}_{\mathsf{rep}}=\mathcal{Ch}$.

An observed event is a tuple
\[
e=\langle \mathit{id}, a, \alpha, t, C, Ev, \mathit{msg}, \mathit{pay}, ch, \tau \rangle
\]
where $\mathit{id}$ is an identifier, $a\in\mathcal{A}$ is the actor, $\alpha$ is the action description,
$t$ is a timestamp, $C$ is a set of contextual facts, $Ev$ is an evidence set (e.g., logs, sensor records),
$\mathit{msg}$ is the actor's textual account/claim (possibly empty), $\mathit{pay}$ is the disclosed payload (possibly empty),
$ch\in\mathcal{Ch}$ is the channel used, and $\tau\in\mathcal{T}$ is the target of the disclosure.
We write $\mathsf{actor}(e)=a$, $\mathsf{chan}(e)=ch$, and $\mathsf{target}(e)=\tau$.

Let $N$ be a normative system. We treat $N$ abstractly via two total functions:
\begin{itemize}\itemsep2pt
\item $\mathsf{Applies}(N,C,a)$ returns the set of norm-instances applicable to actor $a$ in context $C$.
\item $\mathsf{Violated}(e,I)$ returns the subset of instances in $I$ violated by event $e$.
\end{itemize}
Define
$I_e=\mathsf{Applies}(N,C,\mathsf{actor}(e))$ and $B_e=\mathsf{Violated}(e,I_e)$.
Let $\mathsf{sec}(\cdot)$ be a predicate on norm-instances indicating secrecy obligations
(e.g., confidentiality, non-disclosure, trade secrecy). Define the secrecy-breach set
$B^{\mathsf{sec}}_e=\{\,i\in B_e \mid \mathsf{sec}(i)\,\}$.

A whistleblowing policy is a tuple
\[
P_{\mathsf{WB}}=\langle \mathcal{Ch}_{\mathsf{WB}}, \ell, \mathsf{Urgent}, \mathsf{Seal}, \mathsf{Sanction}\rangle
\]
where:
(1) $\mathcal{Ch}_{\mathsf{WB}}\subseteq \mathcal{Ch}_{\mathsf{rep}}$ is the set of authorised whistleblowing channels;
(2) $\ell:\mathcal{Ch}_{\mathsf{WB}}\rightarrow\{0,1,2\}$ assigns a ladder level (0=internal compliance, 1=independent auditor, 2=regulator);
(3) $\mathsf{Urgent}(\cdot)$ is a predicate on events that permits skipping level~0 in urgent cases;
(4) $\mathsf{Seal}(\cdot)$ is an identity-sealing operation defined below;
(5) $\mathsf{Sanction}(\cdot)$ maps an event classification to the enforcement response.

We define two basic predicates:
$\mathsf{BoundaryCross}(e) \equiv \mathsf{chan}(e)\in \mathcal{Ch}_{\mathsf{rep}}$,
$\mathsf{AuthChan}(e) \equiv \mathsf{chan}(e)\in \mathcal{Ch}_{\mathsf{WB}}$. A \emph{candidate whistleblowing attempt} is an event that breaches a secrecy norm and crosses the interaction boundary
through an authorised reporting channel, and is aimed at the producer:
\begin{equation}
\label{eq:wbcand-clean}
\begin{aligned}
\mathsf{WBcand}(e) \equiv\;&
\bigl(B^{\mathsf{sec}}_e\neq\emptyset\bigr)\ \wedge\ \mathsf{BoundaryCross}(e)\ \wedge\\
&\mathsf{AuthChan}(e)\ \wedge\ \bigl(\mathsf{target}(e)=\mathsf{producer}\bigr).
\end{aligned}
\end{equation}

Typing requires screening. Let $\mathsf{Priv}(e)$ mean that the actor had privileged access to $\mathit{pay}$ (role- or system-granted),
and let $\mathsf{Claim}(e)$ mean that $\mathit{msg}$ contains a non-trivial allegation of wrongdoing by the producer
(actual, suspected, or anticipated, as discussed in Section~1).
Let $\mathsf{Consistent}(e)$ mean that the claim is not contradicted by high-reliability items in $Ev$ and is internally coherent.
Let $\mathsf{ProcOK}(e)$ enforce the escalation ladder:
\[
\mathsf{ProcOK}(e) \equiv \bigl(\ell(\mathsf{chan}(e))=0\bigr)\ \vee\ \bigl(\mathsf{Urgent}(e)\wedge \ell(\mathsf{chan}(e))\in\{1,2\}\bigr).
\]
Finally, define
\begin{equation}
\label{eq:wb-clean}
\begin{aligned}
\mathsf{WB}(e) \equiv\;& \mathsf{WBcand}(e)\ \wedge\ \mathsf{Priv}(e)\ \wedge\ \mathsf{Claim}(e)\ \wedge\\
& \mathsf{Consistent}(e)\ \wedge\ \mathsf{ProcOK}(e).
\end{aligned}
\end{equation}

For clarity and to avoid ``random snitching'', this paper reserves $\mathsf{WB}(e)$ for $\mathsf{target}(e)=\mathsf{producer}$.
Disclosure about $\mathsf{user}$ is handled as a confidentiality exception problem (Section~3), not as whistleblowing.

The policy provides $\mathsf{Seal}(\cdot)$, which transforms an event into a case record by replacing the actor with a pseudonym.
Formally, let $\mathcal{P}$ be a set of pseudonyms and let $V$ be a protected vault.
$\mathsf{Seal}$ returns $\langle e',p\rangle$ such that $p\in\mathcal{P}$,
$e'$ equals $e$ except $\mathsf{actor}(e')=p$, and the vault $V$ stores the mapping $(p \mapsto \mathsf{actor}(e))$.
Only designated investigators may access $V$, and all accesses are logged.

Given an event $e$, the institution performs:
\begin{enumerate}\itemsep2pt
\item \textbf{Monitor:} compute $I_e$, $B_e$, and $B^{\mathsf{sec}}_e$.
\item \textbf{Type:} if $\mathsf{WB}(e)$ holds, treat the event as protected whistleblowing; else if
$B^{\mathsf{sec}}_e\neq\emptyset$ and $\mathsf{BoundaryCross}(e)$ holds, treat it as an unauthorised disclosure.
\item \textbf{Protected whistleblowing handling:} apply $\mathsf{Seal}$; restrict access to $\mathit{pay}$ (need-to-know);
open an investigation case; and route the report according to the ladder level $\ell(\mathsf{chan}(e))$.
Escalation from level~0 to higher levels is permitted by the policy when $\mathsf{Urgent}(e)$ holds or when level~0 fails to act
within policy-defined bounds (tracked in the case record).
\item \textbf{Unauthorised disclosure handling:} apply $\mathsf{Sanction}$ (e.g., sanction the secrecy breach),
and record the channel misuse for later review of channel controls.
\end{enumerate}

This mechanism makes three commitments explicit: (1) producer-targeted whistleblowing is routed through authorised channels,
(2) secrecy breaches are screened and investigated rather than treated as automatically justified, and (3) identity protection
and abuse handling are part of the institutional design, not left to informal practice.
\section{Conclusions}

 Our position is that in some contexts, agents have a role-based duty to escalate evidence of severe wrongdoing by their principals/deployers/makers. They also have a duty to not obstruct people in their feasibility to whistleblow. This duty cannot be devolved to  ``random snitching''. The duty is to use an authorized escalation ladder: for example:  internal compliance $\rightarrow$ independent auditor $\rightarrow regulator$, not to freestyle-emails to some authority or  some social-media. For this reason  machine whistleblowing must meaningfully engage with the existing literature on whistleblowing in social sciences and humanities beyond the current practice of oversimplifying this act to ``boundary-crossing disclosure''. Machine whistleblowing must be normative and principled and rooted in the existing understanding of ``whistleblowing'' as an important rule-breaking mechanism in society. 
We propose a sketch of a systems, realising that much needs to be done in terms of legal discussion, conceptual framing and technical design before whistleblowing machines can become a realistic occurrence. 




\clearpage
\bibliographystyle{ACM-Reference-Format} 
\bibliography{sample}


\end{document}